\def\year{2019}\relax
\documentclass[letterpaper]{article} 
\usepackage{aaai2019}  
\usepackage{times}  
\usepackage{helvet}  
\usepackage{courier}  
\usepackage{url}  
\usepackage{graphicx}  
\usepackage{multirow}
\usepackage{amsthm}
\usepackage{amsfonts}
\usepackage{MnSymbol}
\usepackage{makecell}
\usepackage{arydshln}
\usepackage{microtype}
\usepackage{amsmath}
\usepackage{booktabs}

\newcommand{\ours}{RAVEN}
\newcommand{\ourl}{Recurrent Attended Variation Embedding Network}
\newcommand{\attention}{Gated Modality-mixing Network}
\newcommand{\subword}{Nonverbal Sub-networks}
\newcommand{\augmentation}{Multimodal Shifting}
\begin{document}
%
\title{Words Can Shift: \\ Dynamically Adjusting Word Representations Using Nonverbal Behaviors}

\author{Yansen Wang$^1$, Ying Shen$^2$, Zhun Liu$^2$, Paul Pu Liang$^2$, Amir Zadeh$^2$, Louis-Philippe Morency$^2$ \\
$^1$Department of Computer Science, Tsinghua University\\
$^2$School of Computer Science, Carnegie Mellon University \\
\texttt{ys-wang15@mails.tsinghua.edu.cn;  \{yshen2,zhunl,pliang,abagherz,morency\}@cs.cmu.edu} \\
}
\maketitle

\begin{abstract}
Humans convey their intentions through the usage of both verbal and nonverbal behaviors during face-to-face communication. Speaker intentions often vary dynamically depending on different nonverbal contexts, such as vocal patterns and facial expressions. As a result, when modeling human language, it is essential to not only consider the literal meaning of the words but also the nonverbal contexts in which these words appear. To better model human language, we first model expressive nonverbal representations by analyzing the fine-grained visual and acoustic patterns that occur during word segments. In addition, we seek to capture the dynamic nature of nonverbal intents by shifting word representations based on the accompanying nonverbal behaviors. To this end, we propose the Recurrent Attended Variation Embedding Network (RAVEN) that models the fine-grained structure of nonverbal subword sequences and dynamically shifts word representations based on nonverbal cues. Our proposed model achieves competitive performance on two publicly available datasets for multimodal sentiment analysis and emotion recognition. We also visualize the shifted word representations in different nonverbal contexts and summarize common patterns regarding multimodal variations of word representations.
\end{abstract}

\section{Introduction}

\begin{figure}[!ht]
\centering{
\includegraphics[width=1.0\linewidth]{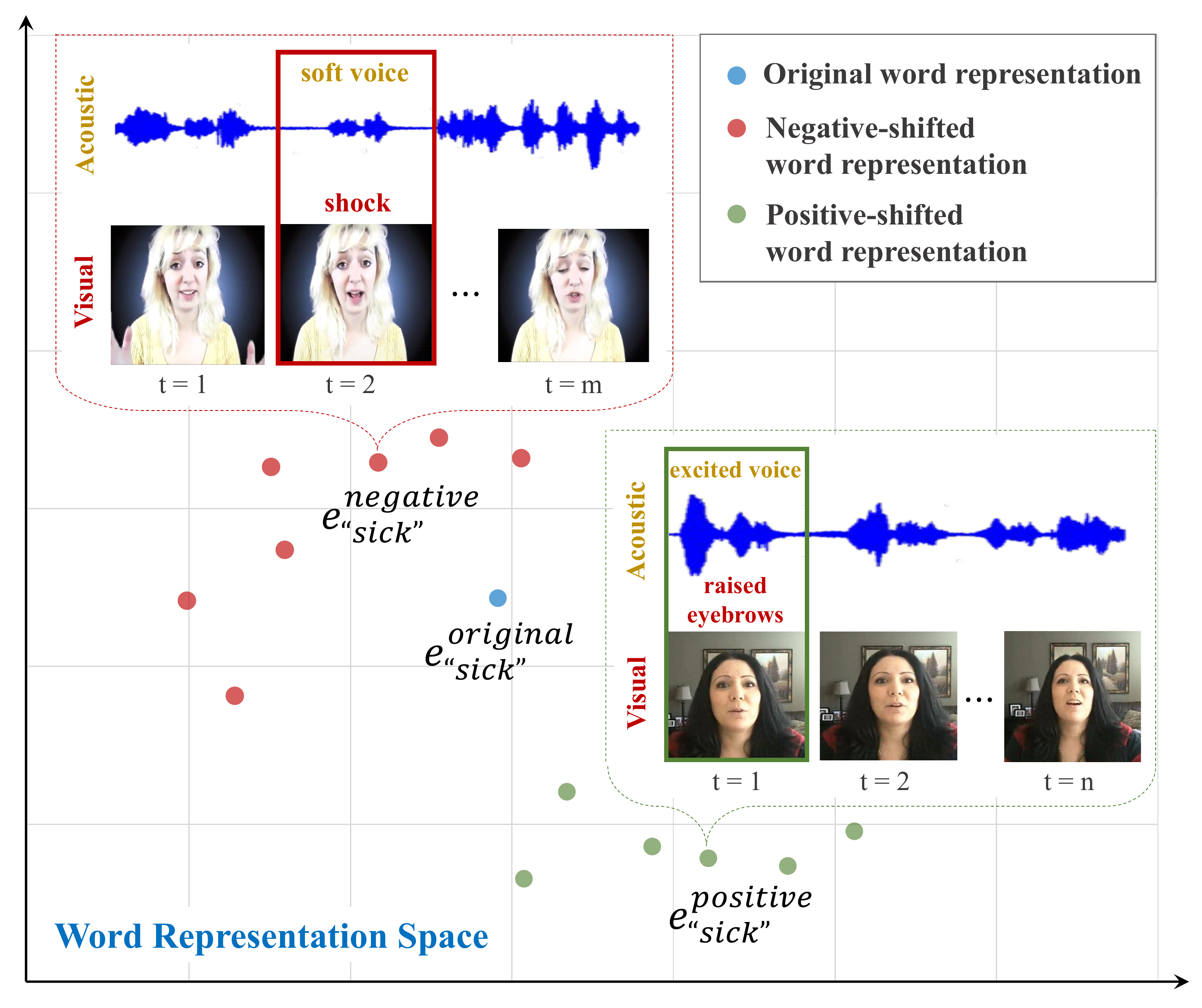}}
\caption{Conceptual figure demonstrating that the word representation of the same underlying word ``sick" can vary conditioned on different co-occurring nonverbal behaviors. The nonverbal context during a word segment is depicted by the sequence of facial expressions and intonations. The speaker who has a relatively soft voice and frowning behaviors at the second time step displays negative sentiment.
}
\label{fig:conceptual}
\end{figure}

Multimodal language communication happens through both verbal and nonverbal channels. The verbal channel of communication conveys intentions through words and sentences while the nonverbal aspect uses gestures and vocal intonations. However, the meaning of words and sentences uttered by the speaker often varies dynamically in different nonverbal contexts. These dynamic behaviors can arise from different sources such as cultural shift or different political backgrounds~\cite{bamler2017dynamic}. In human multimodal language, these dynamic behaviors are often intertwined with their nonverbal contexts \cite{burgoon2016nonverbal}. Intentions conveyed through uttering a sentence can display drastic shifts in intensity and direction, leading to the phenomena that the uttered words exhibit dynamic meanings depending on different nonverbal contexts.

Previous work in modeling human language often utilizes word embeddings pretrained on a large textual corpus to represent the meaning of language. However, these methods are not sufficient for modeling highly dynamic human multimodal language. The example in Figure~\ref{fig:conceptual} demonstrates how the same underlying word can vary in sentiment when paired with different nonverbal cues. Although the two speakers are using the same adjective ``sick" to describe movies, they are conveying different sentiments and remarks by showing opposing facial expressions and intonations. These subtle nonverbal cues contained in the span of the uttered words, including facial expressions, facial landmarks, and acoustic features, are crucial towards determining the exact intent displayed in verbal language. We hypothesize that this ``exact intent" can often be derived from the representation of the uttered words combined with a shift in the embedding space introduced by the accompanying nonverbal cues. In this regard, a dynamic representation for words in particular visual and acoustic background is required.

Modeling the nonverbal contexts concurrent to an uttered word requires fine-grained analysis. This is because the visual and acoustic behaviors often have a much higher temporal frequency than words, leading to a sequence of accompanying visual and acoustic ``subword" units for each uttered word. The structure of these subword sequences is especially important towards the representation of nonverbal dynamics. In addition, modeling subword information has become essential for various tasks in natural language processing~\cite{SCLeM:2017}, including language modeling~\cite{labeau-allauzen:2017:SCLeM,Kim:2016:CNL:3016100.3016285}, learning word representations for different languages~\cite{DBLP:journals/corr/abs-1802-05365,DBLP:conf/acl/OhPBBC18,DBLP:journals/corr/BojanowskiGJM16}, and machine translation~\cite{DBLP:journals/corr/abs-1804-10959,DBLP:journals/corr/SennrichHB15}. However, many of these previous works in understanding and modeling multimodal language has ignored the role of subword analysis. Instead, they summarize the subword information during each word span using the simple averaging strategies~\cite{multistage,lowrank,DBLP:conf/aaai/ZadehLPVCM18b}. While average behaviors may be helpful in modeling global characteristics, it is lacking in its representation capacity to accurately model the structure of nonverbal behaviors at the subword level. This motivates the design of a more expressive model that can accurately capture the fine-grained visual and acoustic patterns that occur in the duration of each word.

To this end, we propose the \ourl \ (RAVEN), a model for human multimodal language that considers the fine-grained structure of nonverbal subword sequences and dynamically shifts the word representations based on these nonverbal cues. In order to verify our hypotheses on the importance of subword analysis as well as the dynamic behaviors of word meanings, we conduct experiments on multimodal sentiment analysis and emotion recognition. Our model shows excellent performance on both tasks. We present visualizations of the shifted word representations to better understand the impact of subword modeling and dynamic shifts on modeling word meaning. Finally, we present ablation studies to analyze the effects of subword modeling and dynamic shifting. We discover that the shifted embeddings learned by \ours \ exhibit meaningful distributional patterns with respect to the sentiment expressed by the speaker.

\section{Related Works}

Previously, much effort has been devoted to building machine learning models that learn from multiple modalities~\cite{ngiam2011multimodal,JMLR:v15:srivastava14b}. However, there has been limited research into modeling the variations of word representations using nonverbal behaviors. To place our work in the context of prior research, we categorize previous works as follows: (1) {subword word representations}, (2) {modeling variations in word representations}, and (3) {multimodal sentiment and emotion recognition}.

\textit{Modeling subword information} has become crucial for various tasks in natural language processing~\cite{SCLeM:2017}. Learning the compositional representations from subwords to words allows models to infer representations for words not in the training vocabulary. This has proved especially useful for machine translation~\cite{DBLP:journals/corr/SennrichHB15}, language modeling~\cite{Kim:2016:CNL:3016100.3016285} and word representation learning~\cite{DBLP:journals/corr/BojanowskiGJM16}. In addition, deep word representations learned via neural models with character convolutions~\cite{NIPS2015_5782} have been found to contain highly transferable language information for downstream tasks such as question answering, textual entailment, sentiment analysis, and natural language inference~\cite{DBLP:journals/corr/abs-1802-05365}.

\textit{Modeling variations in word representations} is an important research area since many words have different meanings when they appear in different contexts. \citeauthor{DBLP:journals/corr/LiJ15a}~\shortcite{DBLP:journals/corr/LiJ15a} propose a probabilistic method based on Bayesian Nonparametric models to learn different word representations for each sense of a word, \citeauthor{DBLP:journals/corr/NguyenNMTP17}~\shortcite{DBLP:journals/corr/NguyenNMTP17} use a Gaussian Mixture Model~\cite{reference/bio/Reynolds09} and \citeauthor{P18-1001}~\shortcite{P18-1001} extend FastText word representations~\cite{DBLP:journals/corr/BojanowskiGJM16} with a Gaussian Mixture Model representation for each word.

Prior work in \textit{multimodal sentiment and emotion recognition} has tackled the problem via multiple approaches: the early fusion method refers to concatenating multimodal data at the input level. While these methods are able to outperform unimodal models~\cite{zadeh2016multimodal} and learn robust representations~\cite{wang2016select}, they have limited capabilities in learning modality-specific interactions and tend to overfit~\cite{xu2013survey}. The late fusion method integrates different modalities at the prediction level. These models are highly modular, and one can build a multimodal model from individual pre-trained unimodal models and fine-tuning on the output layer~\cite{poria2017context}. While such models can also outperform unimodal models~\cite{seq2seq}, they focus mostly on modeling modality-specific interactions rather than cross-modal interactions. Finally, multi-view learning refers to a broader class of methods that perform fusion between the input and prediction levels. Such methods usually perform fusion throughout the multimodal sequence~\cite{rajagopalan2016extending,localglobal}, leading to explicit modeling of both modality-specific and cross-modal interactions at every time step. Currently, the best results are achieved by augmenting this class of models with attention mechanisms~\cite{multistage}, word-level alignment~\cite{factorized}, and more expressive fusion methods~\cite{lowrank}.

\begin{figure*}[t!]
\centering{
\includegraphics[width=1.0\linewidth]{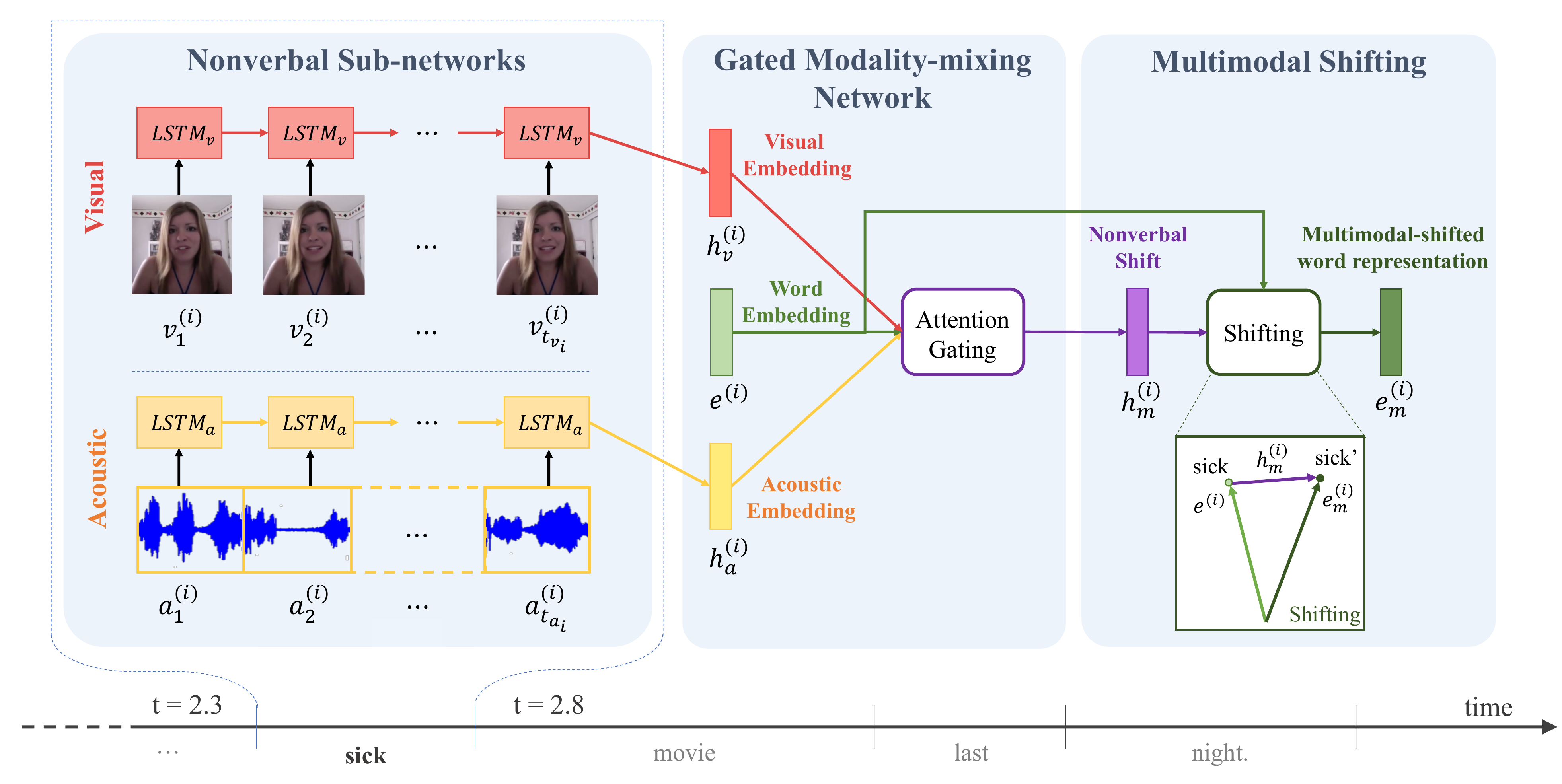}}
\caption{
An illustrative example for \ourl\ (\ours) model: The \ours\ model has three components: (1) \subword, (2) \attention, and (3) \augmentation. For the given word ``sick" in the utterance, the \subword\ first computes the visual and acoustic embedding through modeling the sequence of visual and acoustic features lying in a word-long segment with separate LSTM network. The \attention\ module then infers the nonverbal shift vector as the weighted average over the visual and acoustic embedding based on the original word embedding. The \augmentation\ finally generates the multimodal-shifted word representation by integrating the nonverbal shift vector to the original word embedding. The multimodal-shifted word representation can be then used in the high-level hierarchy to predict sentiments or emotions expressed in the sentence.
}
\label{fig:overview}
\end{figure*}

These previous studies have explored integrating non-verbal behaviors or building word representations with different variations from purely textual data. However, these works do not consider the temporal interactions between the nonverbal modalities that accompany the language modality at the subword level, as well as the contribution of non-verbal behaviors towards the meaning of underlying words. Our proposed method models the nonverbal temporal interactions between the subword units. This is performed by word-level fusion with nonverbal features introducing variations to word representations. In addition, our work can also be seen as an extension of the research performed in modeling multi-sense word representations. We use the accompanying nonverbal behaviors to learn variation vectors that either (1) disambiguate or (2) emphasize the existing word representations for multimodal prediction tasks.

\section{\ourl \ (\ours)}

The goal of our work is to better model multimodal human language by (1) considering subword structure of nonverbal behaviors and (2) learning multimodal-shifted word representations conditioned on the occurring nonverbal behaviors. To achieve this goal, we propose the \ourl\ (\ours).

An overview of the proposed \ours\ model is given in Figure \ref{fig:overview}. Our model consists of three major components: (1) \textit{\subword} model the fine-grained structure of nonverbal behaviors at the subword level by using two separate recurrent neural networks to encode a sequence of visual and acoustic patterns within a word-long segment, and outputs the nonverbal embeddings. (2) \textit{\attention} takes as input the original word embedding as well as the visual and acoustic embedding, and uses an attention gating mechanism to yield the nonverbal shift vector which characterizes how far and in which direction has the meaning of the word changed due to nonverbal context.
(3) \textit{\augmentation} computes the multimodal-shifted word representation by integrating the nonverbal shift vector to the original word embedding. The following subsections discuss the details of these three components of our \ours\ model.


\subsection{\subword}

To better model the subword structure of nonverbal behaviors, the proposed \subword \ operate on the visual and acoustic subword units carried alongside each word. This yields the visual and acoustic embeddings. These output embeddings are illustrated in Figure \ref{fig:overview}.

Formally, we begin with a segment of multimodal language $\mathbf{L}$ denoting the sequence of uttered words. For the span of the $i$th word denoted as $\mathbf{L}^{(i)}$, we have two accompanying sequences from the visual and acoustic modalities: $\mathbf{V}^{(i)} = [v^{(i)}_1, v^{(i)}_2, \cdots, v^{(i)}_{t_{v_i}}]$, $\mathbf{A}^{(i)} = [a^{(i)}_1, a^{(i)}_2, \cdots, a^{(i)}_{t_{a_i}}]$. These are temporal sequences of visual and acoustic frames, to which we refer as the visual and acoustic subword units. To model the temporal sequences of sub-word information coming from each modality and compute the nonverbal embeddings, we use Long-short Term Memory (LSTM) \cite{hochreiter1997long} networks. LSTMs have been successfully used in modeling temporal data in both computer vision~\cite{ullah2018action} and acoustic signal processing \cite{hughes2013recurrent}.

The modality-specific LSTMs are applied to the sub-word sequences for each word $\mathbf{L}^{(i)}, i=1,\cdots, n$. For the $i$-th word $\mathbf{L}^{(i)}$ in the language modality, two LSTMs are applied separately for its underlying visual and acoustic sequences:
\begin{align}
    \mathbf{h}^{(i)}_v = \mathtt{LSTM}_v(\mathbf{V}^{(i)}) \\
    \mathbf{h}^{(i)}_a = \mathtt{LSTM}_a(\mathbf{A}^{(i)})
\end{align}
where $\mathbf{h}^{(i)}_v$ and $\mathbf{h}^{(i)}_a$ refer to the final states of the visual and acoustic LSTMs. We call these final states visual and acoustic embedding, respectively.

\subsection{\attention}

Our \attention\ component computes the nonverbal shift vector by learning a non-linear combination between the visual and acoustic embedding using an attention gating mechanism. Our key insight is that depending on the information in visual and acoustic modalities as well as the word that is being uttered, the relative importance of the visual and acoustic embedding may differ. For example, the visual modality may demonstrate a high activation of facial muscles showing shock while the tone in speech may be uninformative. To handle these dynamic dependencies, we propose a gating mechanism that controls the importance of each visual and acoustic embedding.

In order for the model to control how strong a modality's influence is, we use modality-specific influence gates to model the intensity of the influence. To be more concrete, for word $\mathbf{L}^{(i)}$, given the original word representation $\mathbf{e}^{(i)}$, we concatenate $\mathbf{e}^{(i)}$ with the visual and acoustic embedding $\mathbf{h}^{(i)}_v$ and $\mathbf{h}^{(i)}_a$ respectively and then use the concatenated vectors as the inputs of the visual and acoustic gate $w_v^{(i)}$ and $w_a^{(i)}$:
\begin{align}
    w^{(i)}_v = \sigma (\mathbf{W}_{hv} [\mathbf{h}^{(i)}_v;\mathbf{e}^{(i)}] + b_v) \\
    w^{(i)}_a = \sigma (\mathbf{W}_{ha} [\mathbf{h}^{(i)}_a;\mathbf{e}^{(i)}] + b_a)
\end{align}
where $[;]$ denotes the operation of vector concatenation. $\mathbf{W}_{hv}$ and $\mathbf{W}_{ha}$ are weight vectors for the visual and acoustic gates and $b_v$ and $b_a$ are scalar biases. The sigmoid function $\sigma (x)$ is defined as $\sigma(x) = \frac{1}{1+e^{-x}}, x \in \mathbb{R}$.



Then a nonverbal shift vector is calculated by fusing the visual and acoustic embeddings multiplied by the visual and acoustic gates. Specifically, for a word $\mathbf{L}^{(i)}$, the nonverbal shift vector $\mathbf{h}^{(i)}_m$ is calculated as follows:
\begin{align}\label{var}
    \mathbf{h}^{(i)}_m = w_v^{(i)} \cdot (\mathbf{W}_{v}& \mathbf{h}^{(i)}_v) + w^{(i)}_a \cdot (\mathbf{W}_{a} \mathbf{h}^{(i)}_a) + \mathbf{b}^{(i)}_h
\end{align}
where $\mathbf{W}_{v}$ and $\mathbf{W}_{a}$ are weight matrices for the visual and acoustic embedding and $\mathbf{b}^{(i)}_h$ is the bias vector.


\subsection{\augmentation}

The \augmentation\ component learns to dynamically shift the word representations by integrating the nonverbal shift vector $\mathbf{h}^{(i)}_m$ into the original word embedding. Concretely, the multimodal-shifted word representation for word $\mathbf{L}^{(i)}$ is given by:
\begin{align}\label{embed}
    \mathbf{e}^{(i)}_m &= \mathbf{e}^{(i)} + \alpha \mathbf{h}^{(i)}_m \\
    \alpha &= \min \bigg(\frac{||\mathbf{e}^{(i)}||_2}{||\mathbf{h}^{(i)}_m||_2} \beta, 1 \bigg)
\end{align}
where $\beta$ is a threshold hyper-parameter which can be determined by cross-validation on a validation set.

In order to ensure the magnitude of the nonverbal shift vector $\mathbf{h}^{(i)}_m$ is not too large as compared to the original word embedding $\mathbf{e}^{(i)}$, we apply a scaling factor $\alpha$ to constrain the magnitude of the nonverbal shift vector to be within a desirable range. At the same time, the scaling factor maintains the direction of the shift vector.

By applying the same method for every word in $\mathbf{L}$, we can transform the original sequence triplet $(\mathbf{L}, \mathbf{V}, \mathbf{A})$ into one sequence of multimodal-shifted representations $\mathbf{E} = [\mathbf{e}^{(1)}_m, \mathbf{e}^{(2)}_m, \cdots, \mathbf{e}^{(n)}_m]$. The new sequence $\mathbf{E}$ now corresponds to a shifted version of the original sequence of word representations $\mathbf{L}$ fused with information from its accompanying nonverbal contexts.

This sequence of multimodal-shifted word representations is then used in the high-level hierarchy to predict sentiments or emotions expressed in the utterance. We can use a simple word-level LSTM to encode a sequence of the multimodal-shifted word representations into an utterance-level multimodal representation $\mathbf{h}$. This multimodal representation can then be used for downstream tasks:

\begin{align}
    \mathbf{h} &= \mathtt{LSTM}_e(\mathbf{E})
\end{align}


For concrete tasks, the representation $\mathbf{h}$ is passed into a fully-connected layer to produce an output that fits the task. The various components of \ours \ are trained end-to-end together using gradient descent.

\section{Experiments}

In this section, we describe the experiments designed to evaluate our {\ours} model. We start by introducing the tasks and datasets and then move on to the feature extraction scheme. \footnote{The codes are available at \url{https://github.com/victorywys/RAVEN}.}

\subsection{Datasets}
To evaluate our approach, we use two multimodal datasets involving tri-modal human communications: CMU-MOSI~\cite{zadeh2016multimodal} and IEMOCAP~\cite{Busso2008IEMOCAP:Interactiveemotionaldyadic}, for multimodal sentiment analysis and emotion recognition tasks, respectively.

\textit{Multimodal Sentiment Analysis}: we first evaluate our approach for multimodal sentiment analysis. For this task, we choose the {CMU-MOSI} dataset. It comprises 2199 short video segments excerpted from 93 Youtube movie review videos and has real-valued sentiment intensity annotations from $[-3, +3]$. Negative values indicate negative sentiments and vice versa.

\textit{Multimodal Emotion Recognition}: we investigate the performance of our model under a different, dyadic conversational environment for emotion recognition. The {IEMOCAP} dataset we use for this task contains 151 videos about dyadic interactions, where professional actors are required to perform scripted scenes that elicit specific emotions. Annotations for 9 different emotions are present (angry, excited, fear, sad, surprised, frustrated, happy, disappointed and neutral).


\textit{Evaluation Metrics}: since the multimodal sentiment analysis task can be formulated as a regression problem, we evaluate the performance in terms of Mean-absolute Error (MAE) as well as the correlation of model predictions with true labels. On top of that, we also follow the convention of the CMU-MOSI dataset, and threshold the regression values to obtain a categorical output and evaluate the performance in terms of classification accuracy. As for the multimodal emotion recognition, the labels for every emotion are binary so we evaluate it in terms of accuracy and F1 score.

\subsection{Unimodal Feature Representations}
Following prior practice \cite{lowrank,multistage,P18-1207}, we adopted the same feature extraction scheme for language, visual and acoustic modalities.

\textit{Language Features}: we use the GloVe vectors from~\cite{pennington2014glove}. In our experiments, we used the 300-dimensional version trained on 840B tokens\footnote{https://nlp.stanford.edu/projects/glove/}.

\textit{Visual Features}: given that the two multimodal tasks all include a video clip with the speakers' facial expressions, we employ the facial expression analysis toolkit FACET\footnote{https://imotions.com/} as our visual feature extractor. It extracts features including facial landmarks, action units, gaze tracking, head pose and HOG features at the frequency of 30Hz.

\textit{Acoustic Features}: we use the COVAREP~\cite{degottex2014covarep} acoustic analysis framework for feature extraction. It includes 74 features for pitch tracking, speech polarity, glottal closure instants, spectral envelope. These features are extracted at the frequency of 100Hz.

\subsection{Baseline Models}
Our proposed {\ourl} \ ({\ours}) is compared to the following baselines and state-of-the-art models in multimodal sentiment analysis and emotion recognition.

\textit{{Support Vector Machines}} (SVMs)~\cite{cortes1995support} are widely used non-neural classifiers. This baseline is trained on the concatenated multimodal features for classification or regression tasks~\cite{perez2013utterance,park2014computational,zadeh2016multimodal}.

\textit{{Deep Fusion}} (DF)~\cite{nojavanasghari2016deep} performs late fusion by training one deep neural model for each modality and then combining the output of each modality network with a joint neural network.

\textit{{Bidirectional Contextual LSTM}} (BC-LSTM)~\cite{poria2017context} performs context-dependent fusion of multimodal data.

\textit{{Multi-View LSTM}} (MV-LSTM)~\cite{rajagopalan2016extending} partitions the memory cell and the gates inside an LSTM corresponding to multiple modalities in order to capture both modality-specific and cross-modal interactions.

\textit{{Multi-attention Recurrent Network}} (MARN)~\cite{DBLP:conf/aaai/ZadehLPVCM18b} explicitly models interactions between modalities through time using a neural component called the Multi-attention Block (MAB) and storing them in the hybrid memory called the Long-short Term Hybrid Memory (LSTHM).

\textit{{Memory Fusion Network}} (MFN)~\cite{DBLP:conf/aaai/ZadehLMPCM18a} continuously models the view-specific and cross-view interactions through time with a special attention mechanism and summarized through time with a Multi-view Gated Memory.

\textit{{Recurrent Multistage Fusion Network}} (RMFN)~\cite{multistage} decomposes the fusion problem into multiple stages to model temporal, intra-modal and cross-modal interactions.

\textit{{Low-rank Multimodal Fusion}} (LMF) model~\cite{lowrank} learns both modality-specific and cross-modal interactions by performing efficient multimodal fusion with modality-specific low-rank factors.

\section{Results and Discussion}





In this section, we present results for the aforementioned experiments and compare our performance with state-of-the-art models. We also visualize the multimodal-shifted representations and show that they form interpretable patterns. Finally, to gain a better understanding of the importance of subword analysis and multimodal shift, we perform ablation studies on our model by progressively removing \subword\ and \augmentation\ from our model, and find that the presence of both is critical for good performance.

\subsection{Comparison with the State of the Art}

We present our results on the multimodal datasets in Tables~\ref{table:mosi} and~\ref{table:iemocap}. Our model shows competitive performance when compared with state-of-the-art models across multiple metrics and tasks. Note that our model uses only a simple LSTM for making predictions. This model can easily be enhanced with more advanced modules such as temporal attention.

\textit{Multimodal Sentiment Analysis}: On the multimodal sentiment prediction task, {\ours} achieves comparable performance to previous state-of-the-art models as shown in Table~\ref{table:mosi}. Note the multiclass accuracy \textit{Acc-7} is calculated by mapping the range of continuous sentiment values into a set of intervals that are used as discrete classes.

\textit{Multimodal Emotion Recognition}: On the multimodal emotion recognition task, the performance of our model is also competitive compared to previous ones across all emotions on both the accuracy and F1 score.

\newcolumntype{K}[1]{> {\centering\arraybackslash}p{#1}}
\begin{table}[t!]
\fontsize{7.5}{10}\selectfont
\centering
\setlength\tabcolsep{13.5pt}
\begin{tabular}{l : *{4}{K{1.1cm}}}
\Xhline{0.5\arrayrulewidth}
Dataset & \multicolumn{3}{c}{\textbf{CMU-MOSI}} \\
Metric        & MAE & Corr & Acc-2 \\
\Xhline{0.5\arrayrulewidth}
SVM & 1.864 & 0.057 & 50.2\\
DF &   1.143     &  0.518   & 72.3 \\
BC-LSTM & 1.079 &  0.581 & 73.9 \\
MV-LSTM    & 1.019 &  0.601 & 73.9 \\
MARN  & 0.968  & 0.625 & 77.1    \\
MFN      &  0.965    & 0.632 & 77.4$^\ddagger$    \\
RMFN    &  0.922$^\ddagger$ & 0.681$^\dagger$ & 78.4$^\star$\\
LMF & 0.912$^\star$ & 0.668$^\ddagger$ & 76.4\\
\Xhline{0.5\arrayrulewidth}
{\ours}    & 0.915$^\dagger$    & 0.691$^\star$ & 78.0$^\dagger$  \\
\Xhline{0.5\arrayrulewidth}
\end{tabular}
\caption{Sentiment prediction results on the CMU-MOSI test set using multimodal methods.
The best three results are noted with $^\star$, $^\dagger$ and $^\ddagger$ successively.
}
\label{table:mosi}
\end{table}

\begin{table}[th!]
\fontsize{7.5}{10}\selectfont
\setlength\tabcolsep{2.8pt}
\begin{tabular}{l : *{16}{K{0.673cm}}}
\Xhline{0.5\arrayrulewidth}
Dataset & \multicolumn{8}{c}{\textbf{IEMOCAP Emotions}} \\
Task & \multicolumn{2}{c}{Happy} & \multicolumn{2}{c}{Sad} & \multicolumn{2}{c}{Angry} & \multicolumn{2}{c}{Neutral} \\
Metric   & Acc-2 & F1 & Acc-2 & F1 & Acc-2 & F1 & Acc-2 & F1 \\
\Xhline{0.5\arrayrulewidth}
SVM             & 86.1 & 81.5 & 81.1 & 78.8 & 82.5 & 82.4 & 65.2 & 64.9 \\
DF               & 86.0 & 81.0 & 81.8 & 81.2 & 75.8 & 65.4 & 59.1 & 44.0 \\
BC-LSTM            & 84.9 & 81.7 & 83.2 & 81.7 & 83.5 & 84.2 & 67.5 & 64.1 \\
MV-LSTM           & 85.9 & 81.3 & 80.4 & 74.0 & 85.1$^\ddagger$ & 84.3$^\ddagger$ & 67.0 & 66.7 \\
MARN            & 86.7 & 83.6 & 82.0 & 81.2 & 84.6 & 84.2 & 66.8 & 65.9 \\
MFN                & 86.5 & 84.0 & 83.5$^\dagger$ & 82.1 & 85.0 & 83.7 & 69.6$^\ddagger$ & 69.2$^\ddagger$ \\
RMFN            & 87.5$^\star$ & 85.8$^\star$ & 82.9 & 85.1$^\dagger$ & 84.6 & 84.2 & 69.5 & 69.1 \\
LMF     & 87.3$^\dagger$ & 85.8$^\star$ & 86.2$^\star$ & 85.9$^\star$ & 89.0$^\star$ & 89.0$^\star$ & 72.4$^\star$ & 71.7$^\star$ \\
\Xhline{0.5\arrayrulewidth}
{\ours}  & 87.3$^\dagger$ & 85.8$^\star$ & 83.4$^\ddagger$ & 83.1$^\ddagger$ & 87.3$^\dagger$ & 86.7$^\dagger$ & 69.7$^\dagger$ & 69.3$^\dagger$ \\
\Xhline{0.5\arrayrulewidth}
\end{tabular}
\caption{Emotion recognition results on IEMOCAP test set using multimodal methods. The best three results are noted with $^\star$, $^\dagger$ and $^\ddagger$ successively.
}
\label{table:iemocap}
\end{table}

\subsection{Multimodal Representations in Different Nonverbal Contexts}

\begin{figure*}[t!]
\centering{
\includegraphics[width=1.0\linewidth]{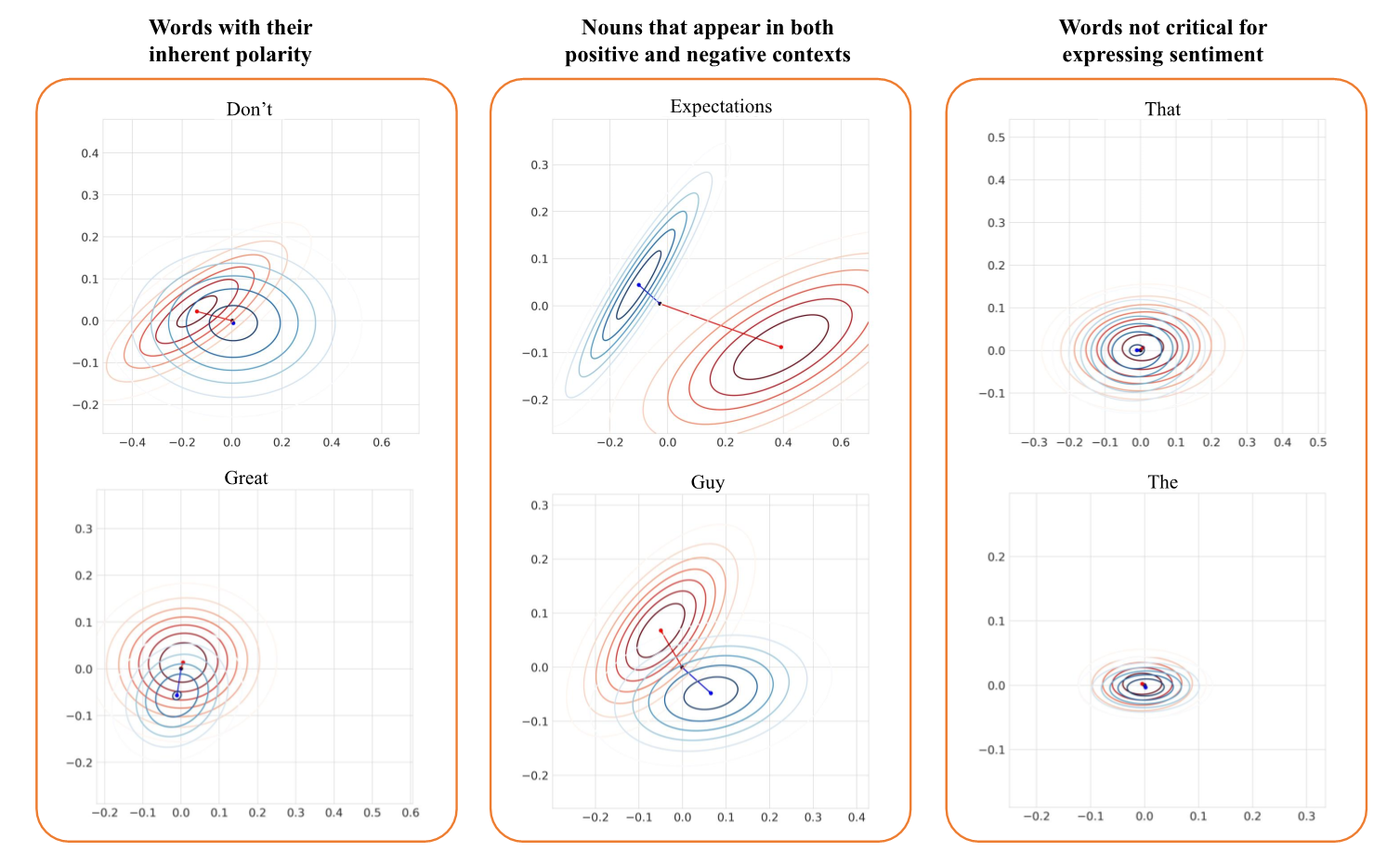}}
\caption{The Gaussian contours of shifted embeddings in 2-dimensional space. Three types of patterns observed in the distribution of all instances of the same word type: words with their inherent polarity will need a drastic variation to convey opposite sentiment; nouns that can appear in both positive and negative contexts will have large variations in both cases; words not critical for expressing sentiment minimal variations in both positive and negative contexts and the distribution of positive/negative instances significantly overlap.}
\label{fig:multivar_macro}
\end{figure*}

As our model learns shifted representations by integrating each word with its accompanying nonverbal contexts, every instance of the same word will have a different multimodal-shifted representation. We observe that the shifts across all instances of the same word often exhibit consistent patterns. Using the CMU-MOSI dataset, we visualize the distribution of shifted word representations that belong to the same word. These visualizations are shown in Figure~\ref{fig:multivar_macro}. We begin by projecting each word representation into 2-dimensional space using PCA~\cite{jolliffe2011principal}. For each word, we plot Gaussian contours for the occurrences in positive-sentiment contexts and the occurrences in negative-sentiment contexts individually. Finally, we plot the centroid of all occurrences as well as the centroids of the subset in positive/negative contexts. To highlight the relative positions of these centroids, we add blue and red arrows starting from that overall centroid and pointing towards the positive and negative centroids. We discover that the variations of different words can be categorized into the following three different patterns depending on their roles in expressing sentiment in a multimodal context:

(1) For \textit{words with their inherent polarity}, their instances in the opposite sentiment context often have strong variations that pull them away from the overall centroid. On the other hand, their instances in their default sentiment context usually experience minimal variations and are close to the overall centroid. In Figure~\ref{fig:multivar_macro}, the word ``great" has an overall centroid that is very close to its positive centroid, while its negative centroid is quite far from both overall and positive centroids.

(2) For \textit{nouns that appear in both positive and negative contexts}, both of their positive and negative centroids are quite far away from the overall centroid, and their positive and negative instances usually occupy different half-planes. While such nouns often refer to entities without obvious polarity in sentiment, our model learns to ``polarize" these representations based on the accompanying multimodal context. For example, the noun ``guy" is frequently used for addressing both good and bad actors, and \ours \ is able to shift them accordingly in the word embedding space towards two different directions (Figure~\ref{fig:multivar_macro}).

(3) For \textit{words that are not critical in conveying sentiment} (e.g. stop words), their average variations under both positive and negative contexts are minimal. This results in their positive, negative, and overall centroids all lying close to each other. Two example words that fall under this category are ``that" and ``the" with their centroids shown in Figure~\ref{fig:multivar_macro}.

These patterns show that {\ours} is able to learn meaningful and consistent shifts for word representations to capture their dynamically changing meanings.


\subsection{Ablation studies}
{\ours} consists of three main components for performing multimodal fusion: {\subword}, {\attention} and {\augmentation}. Among these modules, {\subword} and {\augmentation} are explicitly designed to model the subtle structures in non-verbal behaviors and to introduce dynamic variations to the underlying word representations. In order to demonstrate the necessity of these components in modeling multimodal language, we conducted several ablation studies to examine the impact of each component. We start with our full model and progressively remove different components. The different versions of the model are explained as follows:

\textit{{\ours}}: our proposed model that models subword dynamics and dynamically shifts word embeddings.

\textit{{\ours} w/o SUB}: our model without the {\subword}. In this case, the visual and acoustic sequences are averaged into a vector representation, hence the capability of subword modeling is disabled.

\textit{{\ours} w/o SHIFT}: our model without {\augmentation}. Visual and acoustic representations are concatenated with the word embedding before being fed to downstream networks. While this also generates a representation associated with the underlying word, it is closer to a multimodal representation projected into a different space. This does not guarantee that the new representation is a dynamically-varied embedding in the original word embedding space.

\textit{{\ours} w/o SUB\&SHIFT}: our model with both {\subword} and {\augmentation} removed. This leads to a simple early-fusion model where the visual and acoustic sequences are averaged into word-level representations and concatenated with the word embeddings. It loses both the capabilities of modeling subword structures and creating dynamically-adjusted word embeddings.

Table \ref{table:ablation_mosi} shows the results of ablation studies using several different variants of our model. The results show that both {\subword} and {\augmentation} components are necessary for achieving state-of-the-art performance. This further implies that in scenarios where the visual and acoustic modalities are sequences extracted at a higher frequency, the crude averaging method for sub-sampling them to the same frequency of words does hurt performance. Another observation is that given neural networks are universal function approximators~\cite{csaji2001approximation}, the early-fusion model, in theory, is the most flexible model. Yet in practice, our model improves upon the early-fusion model. This implies that our model does successfully capture underlying structures of human multimodal language.

\begin{table}[t!]
\fontsize{7.5}{10}\selectfont
\setlength\tabcolsep{7pt}
\begin{tabular}{l : *{4}{K{1.0cm}}}
\Xhline{0.5\arrayrulewidth}
Dataset & \multicolumn{3}{c}{\textbf{CMU-MOSI}}  \\
Metric    & MAE & Corr & Acc-2  \\
\Xhline{0.5\arrayrulewidth}
{\ours} &  \textbf{0.915} & \textbf{0.691} & \textbf{78.0} &\\
{\ours} w/o SHIFT &  0.954 & 0.666 & 77.7 &\\
{\ours} w/o SUB &  0.934 & 0.652 & 73.9\\
{\ours} w/o SUB\&SHIFT &  1.423 & 0.116 & 50.6 & \\
\Xhline{0.5\arrayrulewidth}
\end{tabular}
\caption{Ablation studies on CMU-MOSI dataset. The complete \ours \ that models subword dynamics and word shifts works best.
}
\label{table:ablation_mosi}
\end{table}

\section{Conclusion}
In this paper, we presented the \ourl\ (\ours). \ours \ models the fine-grained structure of nonverbal behaviors at the subword level and builds multimodal-shifted word representations that dynamically captures the variations in different nonverbal contexts. \ours \ achieves competitive results on well-established tasks in multimodal language including sentiment analysis and emotion recognition. Furthermore, we demonstrate the importance of both subword analysis and dynamic shifts in achieving improved performance via ablation studies on different components of our model. Finally, we also visualize the shifted word representations in different nonverbal contexts and summarize several common patterns regarding multimodal variations of word representations. This illustrates that our model successfully captures meaningful dynamic shifts in the word representation space given nonverbal contexts. For future work, we will explore the effect of dynamic word representations towards other multimodal tasks  involving language and speech (prosody), videos with multiple speakers (diarization), and combinations of static and temporal data (i.e. image captioning).

\section{Acknowledgments}

This material is based upon work partially supported by the National Science Foundation (Award \#1833355) and Oculus VR. Any opinions, findings, conclusions or recommendations expressed in this material are those of the author(s) and do not necessarily reflect the views of National Science Foundation or Oculus VR. No official endorsement should be inferred. We also thank the anonymous reviewers for useful feedback.

\footnotesize
\bibliography{aaai2019}
\bibliographystyle{aaai2019}

\end{document}